\documentclass{svproc}
\usepackage{amssymb}
\usepackage{amsmath}
\usepackage[disable]{todonotes}
\usepackage{siunitx}
\usepackage{url}

\begin{document}
\mainmatter

\title{
A Comparison of Policy Search in Joint Space and Cartesian Space for Refinement of Skills}
\titlerunning{Comparison of Policy Search in Joint Space and Cartesian Space}
\author{Alexander Fabisch}
\authorrunning{Alexander Fabisch}

\institute{
DFKI GmbH,
Robotics Innovation Center,
Robert-Hooke-Str. 1,
D-28359 Bremen,
Germany,
\email{alexander.fabisch@dfki.de}
}

\maketitle              

\begin{abstract}
Imitation learning is a way to teach robots skills that are demonstrated by
humans.
Transfering skills between these different kinematic structures
seems to be straightforward in Cartesian space.
Because of the correspondence problem, however, the result will most likely
not be identical.
This is why refinement is required, for example, by policy search.
Policy search in Cartesian space is prone to reachability problems when using
conventional inverse kinematic solvers.
We propose a configurable approximate inverse kinematic solver and show that
it can accelerate the refinement process considerably.
We also compare empirically refinement in Cartesian space and refinement in
joint space.
\end{abstract}

\section{Refinement of Demonstrated Skills}
Autonomous, mobile robots with manipulators can be benificial in many
different environments, for example, deep sea \cite{Hildebrandt2008} or
household and factories \cite{Lemburg2011}.
In order to create these robots for real, dynamic environments they have
to be able to learn.
Skill learning frameworks for robots often combine various approaches to
leverage intuitive knowledge from humans \cite{Gutzeit2018}. 
Most of these fall into the categories imitation learning and reinforcement
learning (RL). A standard approach to learning skills is to initialize with
a demonstrated movement and then refine the skill with policy search
\cite{Kober2012,Deisenroth2013,Gutzeit2018}.
Both methods are complementary because on the one hand policy search methods
are often local optimizers that require good initialization and
on the other hand imitation learning usually does not produce a perfect
skill from the start because of the correspondence problem \cite{Gutzeit2018},
that is, different kinematic and dynamic properties of the target system and
the demonstrator result in different outcomes.
We argue that it is sometimes better to do policy search directly in task
space because demonstration and main objectives of the learning problem are
given in task space, for example, end-effector poses in Cartesian space.
When policy search methods are used to learn end-effector
trajectories in task space it often occurs that a trajectory is not
completely in the robot's workspace.
This results in reward landscapes that are difficult to optimize.
We will address this problem with an approximation of inverse kinematics (IK)
and investigate when to use it.

\section{Related Work}
We will present the setting in which our problem is embedded
and give a brief overview of inverse kinematics that are related
to our work.

Transfering skills from humans to robots 
is prone to the correspondence problem.
Consider, for example, a reaching trajectory that is recorded from a human
and transfered to the robot's end-effector. Because the robot probably has
a different hand shape,
it cannot grasp the object with exactly the same end-effector trajectory.
In this case, policy search in Cartesian space can be used to refine the
initial trajectory.
Transfering trajectories directly in joint space is even more difficult
because the segment lengths of a robot arm and the human arm are most likely
different.
Examples for skills that have been demonstrated in task space and transfered
to robots are pancake flipping (learned from kinesthetic teaching;
\cite{Kormushev2010}), peg in hole (learned from  tele-operation;
\cite{Krueger2014}), and serving water (kinesthetic teaching with
a similar arm; \cite{Pastor2009}).
There are several reasons for learning trajectories in task space.
Learning in Cartesian space is often easier when the main objectives are
defined in Cartesian space. For example, learning to grasp an object in
joint space might result in complicated Cartesian trajectories.
When we want to transfer a skill from one robot to another it is easier to
go over end-effector poses instead of joint angles because of different
kinematic structures \cite{Pastor2009} because part of the correspondence
problem is solved by forward and inverse kinematics.
All of the mentioned works use dynamic movement primitives (DMP) as
underlying trajectory representation.
In our work, we will use a DMP based on quaternions \cite{Ude2014}
to generate end-effector poses.


Let $\boldsymbol{q}_t$ be the joint angles of a kinematic chain at time $t$.
The forward kinematics of the chain is given by
$f(\boldsymbol{q}_t) = \boldsymbol{p}_t,$
where $\boldsymbol{p}_t$ denotes the end-effector's pose.
An exact solution to the inverse kinematics (IK) problem would be
$f^{-1}(\boldsymbol{p}_t) = \boldsymbol{q}_t.$
However, $f^{-1}$ is usually not a function. Many joint configurations might
result in the same end-effector pose.
Numerical solutions to inverse kinematics handle redundant kinematic
chains that can have many solutions or complex analytical solutions
efficiently.
Widely used approaches are based on the Jacobian pseudo-inverse
or Jacobian transpose \cite{Nilsson2009}. However, sequential quadratic
programming has been shown to outperform these in terms of
joint limit handling and computation time \cite{Beeson2015}.
Indirect formulations of the form
$$
\arg\min_{\boldsymbol{q}_t \in \mathbb{R}^n}
  (\boldsymbol{q}_{t-1} - \boldsymbol{q}_t)^T (\boldsymbol{q}_{t-1} - \boldsymbol{q}_t), \quad
\text{s.t.} \ g_i(\boldsymbol{q}_t) \leq b_i, \quad i = 1, \ldots, m
$$
where the constraints include the Euclidean distance error, the angular
distance error, and joint limits (e.g. \cite{Kumar2010,Fallon2015}),
have a lower success rate than direct formulations of the form
\begin{eqnarray}
\label{ikproblem1}
\arg\min_{\boldsymbol{q}_t \in \mathbb{R}^n}&&
  d(f(\boldsymbol{q}_t), \boldsymbol{p}^{\text{des}}_t))\\
\label{ikproblem2}
s.t. &&
  g_i(\boldsymbol{q}_t) \leq b_i, \quad i = 1, \ldots, m
\end{eqnarray}
where $d$ is a pose distance metric, $\boldsymbol{p}^{\text{des}}_t$
is the desired end-effector pose, and the inequality constraints only
consist of the joint limits \cite{Beeson2015}.
Few works investigate approximate IK solutions.
However, in cases where the desired end-effector pose cannot be reached
exactly, we would like to have at least the closest possible solution.
Unreachable poses might be a problem of robots that have a low number of
joints \cite{Henning2014} or at the borders of workspaces as this can be
seen for example in visualizations of capability maps \cite{Zacharias2007}.
Traditional methods based on the Jacobian pseudoinverse tend to be
instable and run into local minima \cite{Nilsson2009,Henning2014}.

\section{Configurable Approximate Inverse Kinematics}
A disadvantage of previous approaches is that they do not allow to
configure the approximation.
In the domain of robot skill learning it is often not required to reach
each pose in a trajectory exactly.
Consider the problem of learning to grasp an object. At the
beginning of a reaching movement orientation of the end-effector
is not relevant. It only becomes important at the end.
That is the reason why we develop a configurable IK
solver based on \cite{Beeson2015}.
In the IK formulation in Equations \ref{ikproblem1} and \ref{ikproblem2}
we can use a weighted distance metric of the form
\begin{eqnarray*}
d(\boldsymbol{p}_1, \boldsymbol{p}_2) =&&
  w_{\text{pos}}||\boldsymbol{p}_{1, 1:3} - \boldsymbol{p}_{2, 1:3}||_2^2\\
+&& w_{\text{rot}} \left[ \min(|| \log(\boldsymbol{p}_{1, 3:7} * \overline{\boldsymbol{p}}_{2, 3:7})||_2, 2 \pi - || \log(\boldsymbol{p}_{1, 3:7} * \overline{\boldsymbol{p}}_{2, 3:7})||_2)\right]^2,
\end{eqnarray*}
where $\boldsymbol{p}_{1:3}$ represents the position of the end-effector and
$\boldsymbol{p}_{3:7}$ is a quaternion that represents the orientation of the
end-effector.
It consists of a position distance metric and a rotation distance metric.
There are other distance metrics for rotations that we could use
\cite{Huynh2009}. This allows us to set different weights on position and
rotation because it is often much more important to reach a desired position
than the desired rotation. We could extend this pose metric so that
we can weight each of the six degrees of freedom individually.

\section{Experiments}

\subsection{Methods}
We use DMPs as policies because they are well known and stable trajectory
representations that can be used for imitation and RL. In the following
experiments we will only learn the weights of the DMPs.
Metaparameters are constant.
Many policy search algorithms \cite{HeidrichMeisner2008,Kober2010,Neumann2011,Peters2010,Peters2007,Theodorou2010}
are based on similar principles:
a Gaussian distribution is learned from which we sample policy parameters.
We will use Covariance Matrix Adaption Evolution Strategy (CMA-ES;
\cite{Hansen2001}) to learn the policy
because it has few critical hyperparameters.
Even when those are not set perfectly CMA-ES is still reliable.
The most important hyperparameter is the initial step size $\sigma$.
It determines the width of the initial search distribution. If it is not large
enough, the algorithm will first have to increase the step size for
several generations before it can converge.
If it is set too large, convergence will take longer. For our comparison it
is important to select the correct step size ratio between joint space and
Cartesian space so that the results will not be distorted by a wrong choice of
this parameter.
In our experiments, we use the Kuka LBR IIWA 14 R820 robot arm with 7 DOF.
We determined empirically that the initial step size in joint space must be
2 to 3 times higher than in Cartesian space to achieve similar effects
on the robot's end-effector.
In the first two experiments, the optimum solutions are close to the border of the
workspace so that not every orientation can be reached. In all experiments
we compare learning weights of DMPs in joint space, in Cartesian space with
the proposed approximate IK, and in Cartesian space with an ``exact'' IK.
The term exact throughout this paper means a numerical IK
based on the pseudoinverse of the Jacobian that does not move the end-effector
if no valid solution has been found.
We will execute 30 runs per configuration with different random seeds for
CMA-ES. We use the results to plot learning curves with the mean and
standard error of the maximum reward obtained so far.
Each DMP in our setup has 50 weights per joint space or task space dimension.

\subsection{Problems and Results}
The following paragraphs will summarize the results obtained for several
problems that have different reward surfaces.
Implementations for most experiments are available at
\url{https://github.com/rock-learning/approxik}.

\begin{figure}
\begin{tabular}{ccc}
(a)&(d)&(g)\\
\includegraphics[width=0.38\textwidth]{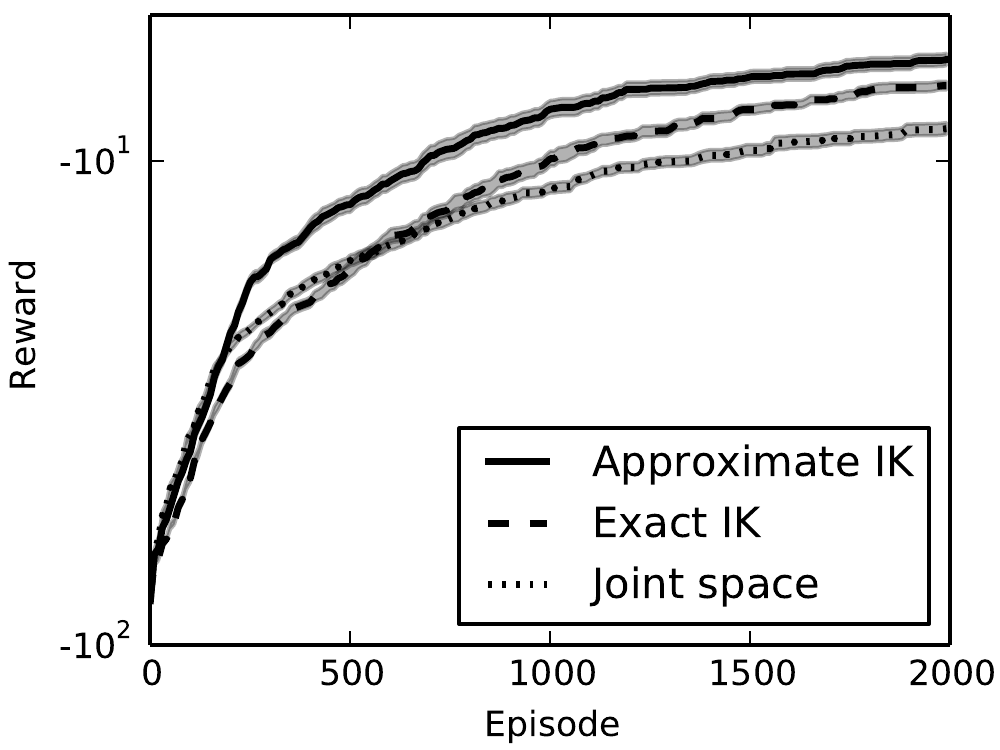}&
\includegraphics[width=0.22\textwidth,clip=true,trim=165 35 165 70]{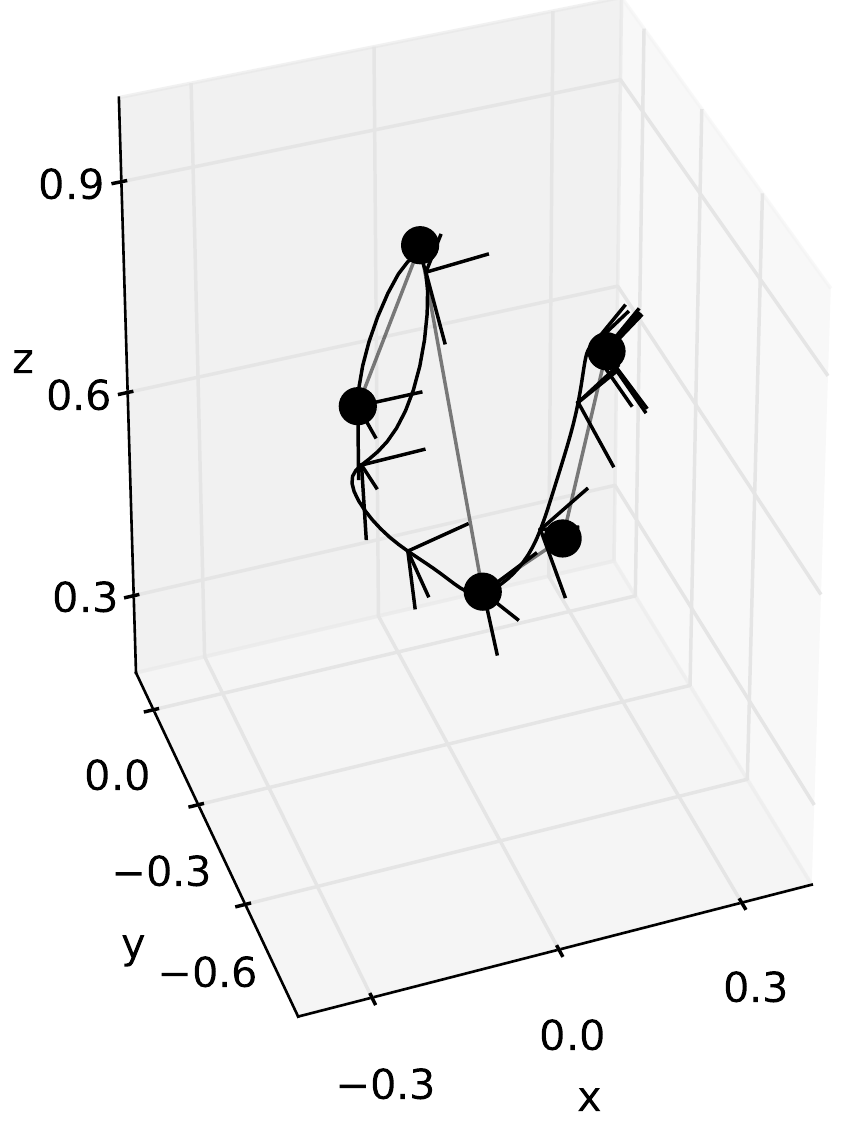}&
\includegraphics[width=0.37\textwidth]{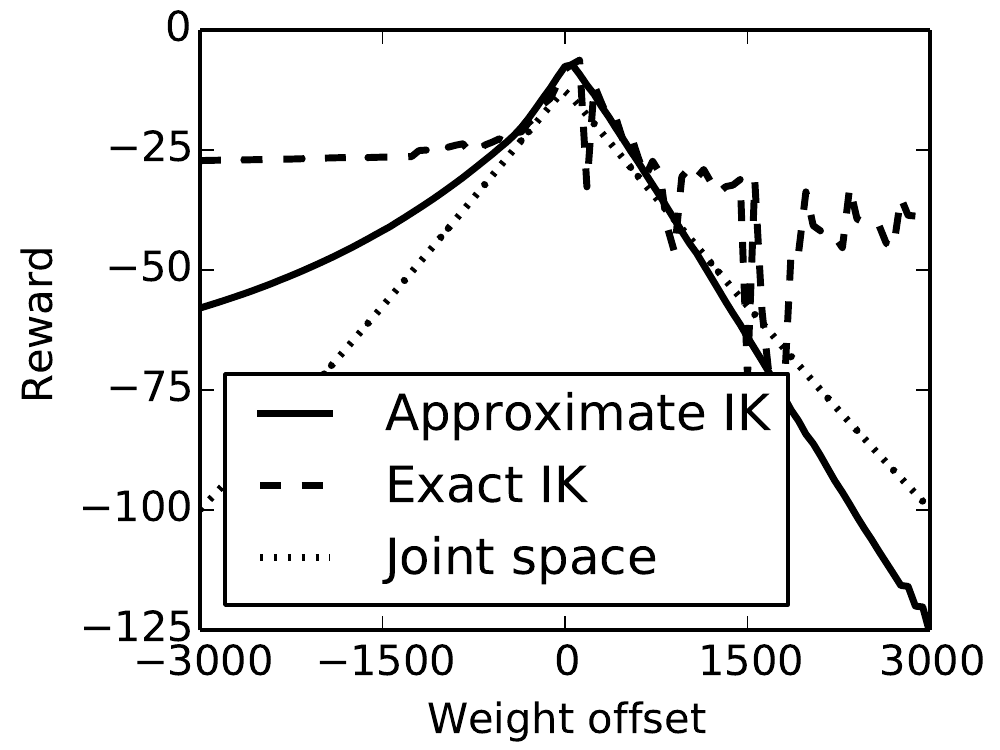}\\
(b)&\multicolumn{2}{c}{(e)}\\
\includegraphics[width=0.38\textwidth]{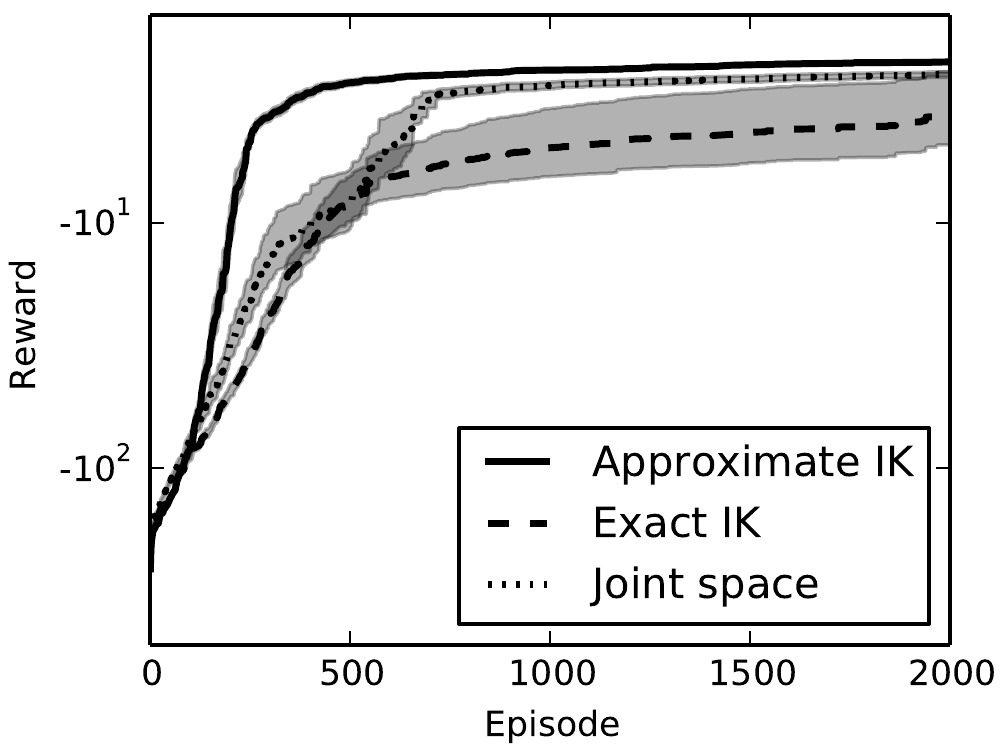}&
\multicolumn{2}{c}{
\includegraphics[width=0.27\textwidth,clip=true,trim=200 80 155 120]{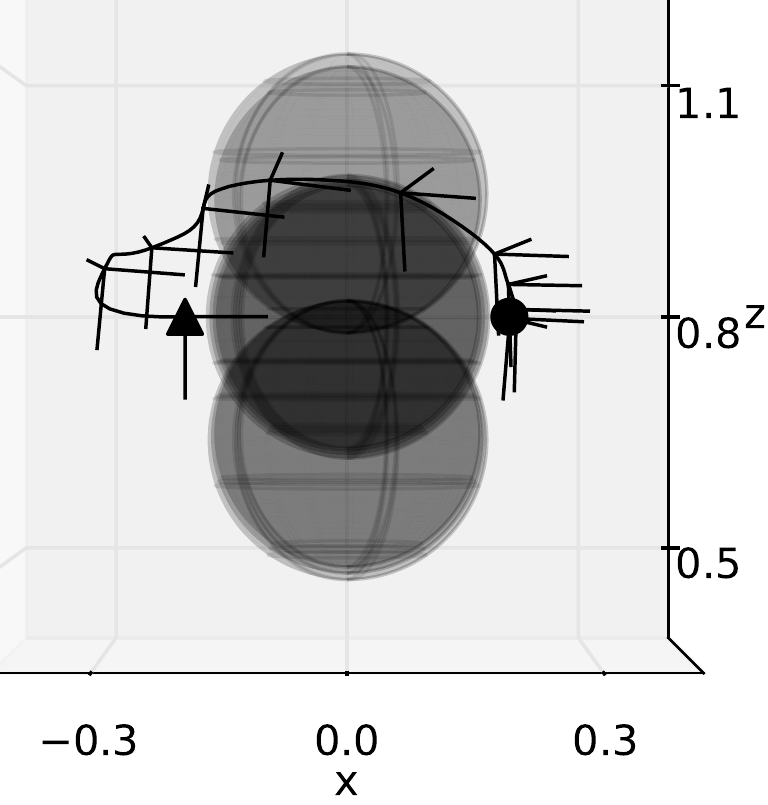}
\includegraphics[width=0.27\textwidth,clip=true,trim=165 35 140 100]{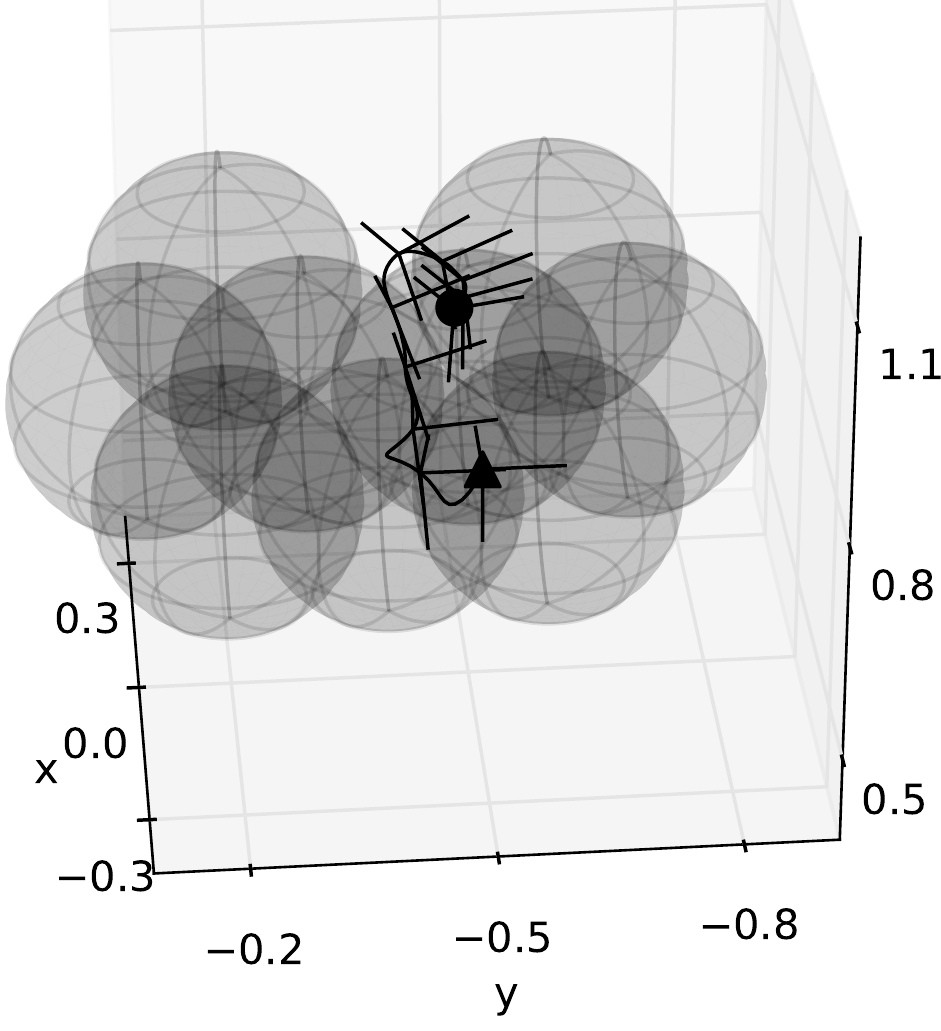}}\\
(c)&\multicolumn{2}{c}{(f)}\\
\includegraphics[width=0.38\textwidth]{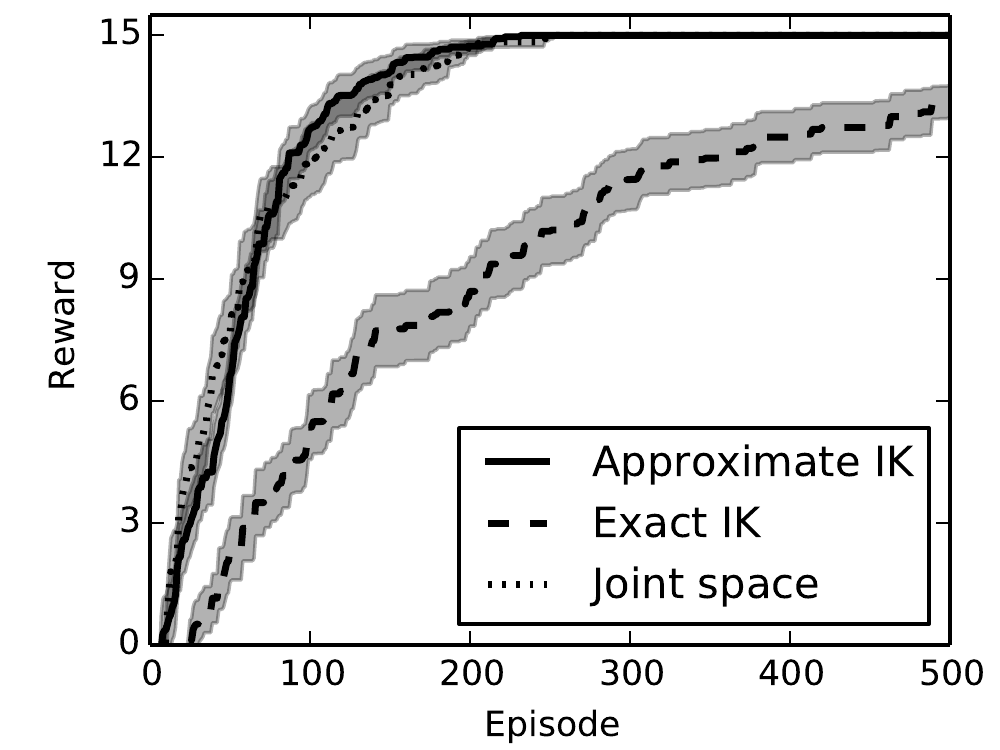}&
\multicolumn{2}{c}{
\includegraphics[width=0.37\textwidth]{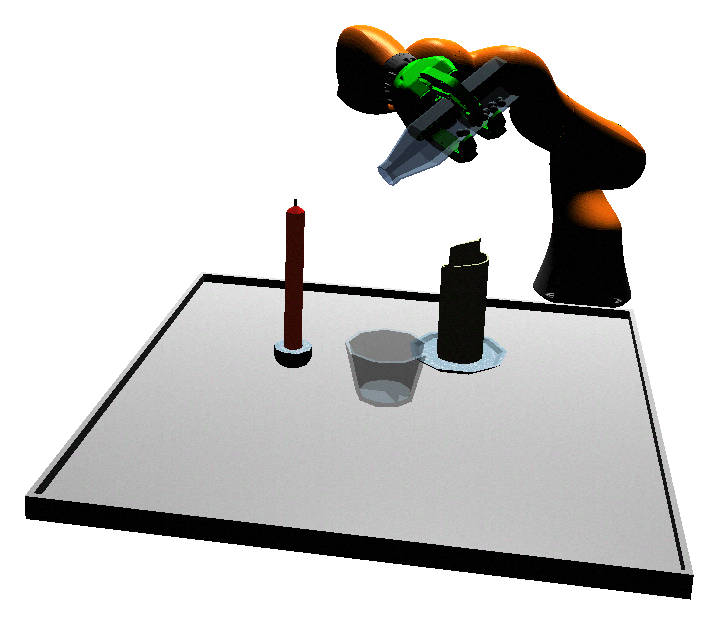}}\\
\multicolumn{3}{c}{(h)}\\
\multicolumn{3}{c}{\includegraphics[width=0.6\textwidth]{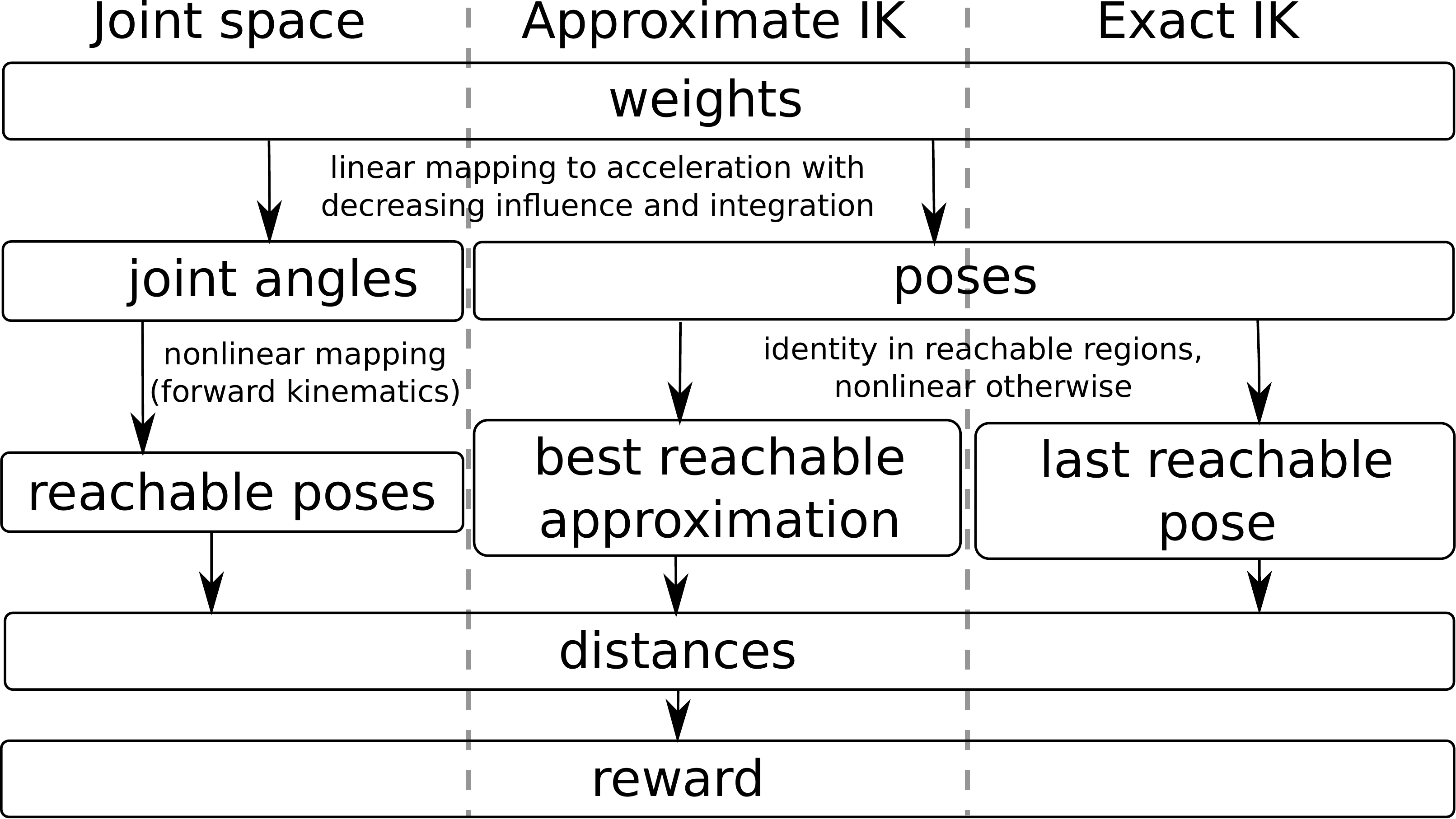}}
\end{tabular}
\centering
\caption{
Learning curves for (a) viapoint, (b) obstacle avoidance, and (c) pouring problem.
Illustrations of environments:
(d) Viapoint: 5 viapoints (circles) have to be reached. A line shows the temporal order.
An additional line indicates an optimized trajectory (rotations are
indicated by small coordinate frames).
(e) Obstacle avoidance: ball-shaped obstacles have to be avoided between start (triangle)
and goal (circle). An example of an optimized trajectory is displayed.
(f) Pouring.
(g) Projection of the reward surface of the viapoint problem on one axis.
(h) Mapping from weights to corresponding reward in the viapoint problem.
For each setup different mappings are involved to compute the reward for
a weight vector.
}
\label{fig:results}
\end{figure}

In the viapoint problem, the end-effector should pass through intermediate positions (viapoints).
A simpler problem has been used to compare various policy search methods
\cite{Kober2010,Peters2006}.
In this version we want the end-effector to pass
through five Cartesian points (see Figure \ref{fig:results} (d))
while minimizing joint velocities and accelerations.
The reward is
$R = -10 \sum_{(t_{\boldsymbol{\nu}},\boldsymbol{\nu})} ||f(\boldsymbol{q}_{t_{\boldsymbol{\nu}}}) - \boldsymbol{\nu}||_2
  - 10^{-3} \sum_{t,j} |\dot{\boldsymbol{q}}_{t,j}|
  - 10^{-5} \sum_{t,j} |\ddot{\boldsymbol{q}}_{t,j}|$,
where $t \in \lbrace 1, \ldots, 101 \rbrace$ represents the step,
$j$ are joint indices and $(t_{\boldsymbol{\nu}}, \boldsymbol{\nu})$
represents a position $\boldsymbol{\nu}$ that has to be reached
at time $t_{\boldsymbol{\nu}}$.
The results are shown in Figure \ref{fig:results} (a).
Learning in Cartesian space is more sample-efficient because the
primary objective is defined in Cartesian space.
The exact IK performs worse because the approximate IK
usually generates more smooth trajectories.
Viapoint problems belong to the most simple class of problems for
policy search methods because there are usually no flat regions or
abrupt changes in the reward surface so that it is easy to determine
the direction of improvement.
We can get a rough impression of the reward surface of the problem
from Figure \ref{fig:results} (g).
To generate a projection of the reward surface on one DMP weight
dimension, we learn a DMP for 1000 episodes, keep the best policy,
modify the 50th weight by adding an offset, and measure the
corresponding reward.
We can see that for both the joint space DMP and the Cartesian DMP
with an approximate IK the reward surface is
smooth with only one maximum. With the exact IK, the reward
surface is rough and abruptly changing in some regions, which
makes learning more difficult for CMA-ES. However, learning in joint
space is worse than learning in Cartesian space because of the
nonlinear mapping from weights to positions in Cartesian space
through the forward kinematics of the robot arm.
This results in more complex interrelations between parameters.



In the obstacle avoidance problem, the end-effector has to avoid several
obstacles represented by spheres.
It is an artificial problem because we only consider end-effector collisions
and not the arm to which it is attached.
The environment is displayed in Figure \ref{fig:results} (e). The reward
is
$R = - 10 \sum_{\boldsymbol{\rho}} p\left(\min_{\boldsymbol{q}_t} ||f(\boldsymbol{q}_t) - \boldsymbol{\rho}||_2\right)
  - 100 ||f(\boldsymbol{q}_T) - \boldsymbol{g}||_2
  - 10^{-2} \sum_{t,j} |\dot{\boldsymbol{q}}_{t,j}|
  - 10^{-5} \sum_{t,j} |\ddot{\boldsymbol{q}}_{t,j}|$,
where $t \in \lbrace 1, \ldots, T \rbrace$, $T=101$ represents the step in time,
$j$ are the joint indices, $\boldsymbol{g}$ is the desired goal position,
and $p(d) = \max(0, 1 - \frac{d}{0.17})$ is greater than zero when the
end-effector is in the vicinity of one of the obstacles.
The result is shown in Figure \ref{fig:results} (b). We see that
learning in Cartesian space is again more sample-efficient because
the primary objective is defined in Cartesian space.
The exact IK performs worse because the approximate
IK and learning in joint space results in more smooth
trajectories.
The difficulty of this task and its reward surface
are similar to those of the viapoint task. The difference
is that it is less smooth because the negative penalty
for being in the vicinity of obstacles abruptly vanishes when the
end-effector is outside of the radius of the spheres. In this case
only the velocity and acceleration penalties guide the search to a
better solution.

In the pouring task, the robot fills a glass with 15 marbles from a bottle
while avoiding nearby obstacles. The setup is displayed in Figure
\ref{fig:results} (f).
The reward function is complex.
For every marble that is inside of the glass we add $1$. For
every marble that is outside of the glass but on the table the negative
squared distance to the center of the glass is given. For every marble that
is still in the bottle we give a reward of $-1$. For collisions with
obstacles we give a reward of $-100$ and abort the episode. Marbles that
fall down the table also finish the episode and a reward of $-1000$ is
given.
Results are displayed in Figure \ref{fig:results} (c).
Learning in joint space and learning in Cartesian space results in
nearly identical learning curves. Using an exact IK
is again worse than using an approximate solution.
The mapping from weights to reward in this problem is complex,
nonlinear, and certainly not smooth. There are abrupt changes in
the reward function when a small change of one weight results in
a marble falling down the table or the arm touching an obstacle.
There are flat regions, for example, when all marbles stay
in the bottle because the arm does not turn the bottle upside down,
or all marbles miss the table. Therefore, the structure of the
reward surface is complex independend of the space in which
we describe DMPs, hence, learning in joint space and learning
in Cartesian space work similarly well.

\subsection{Discussion}
The artificial RL problems that we presented vary in
difficulty, that is, indirection and smoothness of the reward function.
Weights in a DMP are bound to a specific radial basis function.
They only influence one specific joint or pose dimension and only
affect accelerations of the arm locally in time.
They will affect the positions in all following
time steps but with a decreasing influence because the weight of the
learnable forcing term converges to zero.
So it is possible to design a reward function so that the optimization
problem becomes partially separable and, hence, easy to solve.
For example, in the viapoint and obstacle avoidance problems there is a direct
relation between weights of the Cartesian DMP and obtained reward (see Figure
\ref{fig:results} (h)).
With the approximate IK we use the weights to generate a pose trajectory.
An acceleration is computed based on the linear forcing term that includes
the weights and decays exponentially over time.
The acceleration is integrated to obtain a trajectory of
end-effector poses. This trajectory might not be perfectly executable so that
in unreachable regions the pose is mapped to the closest reachable pose.
For each viapoint we compute the distance to the corresponding poses
from the trajectory.
Hence, the mapping from specific weights to the reward in the workspace of
the robot is straightforward and the reward function is partially separable.
Joint space DMPs result in a nonlinear mapping with coupling between dimensions
of the weight space because of the forward kinematics of the robot.
This results in a difficult optimization problem that is not partially
separable any more with probably multiple global and local optima.
The more complex the relation between weights and reward becomes, the
less significant is the difference between learning in joint space and
learning in Cartesian space.
An example is the pouring problem. It has an almost flat reward
surface where the arm collides with an obstacle or a marble falls down.
Small weight changes can make a difference between a
marble staying within the glass and falling down, hence, the reward will
abruptly change in the corresponding regions of the weight space.
The relation between the DMP weights and a successful behavior is
complex, highly nonlinear, and non-separable. This eradicates the
advantage of learning in Cartesian space.

\section{Conclusion}
We develop a configurable, approximate variation of a state-of-the-art
numerical IK solver and show that it accelerates policy
search for movement primitives in Cartesian space in comparison to a
conventional IK solver. Using inverse kinematics can be
regarded as a way to include knowledge about the target system in the
learning process.
Learning in Cartesian space is not always the best option.
It will be advantageous if the main objective has to be solved in
Cartesian space and the reward function is almost separable.
It is not always obvious when this is the case though.
The advantage of learning in Cartesian space vanishes for
more complex, nonlinear, non-separable reward functions.
Policy search works best in the space in which the primary
objectives are defined directly.
Generating smooth trajectories is simpler in joint space.

\section*{Acknowledgements}
{\small
We thank Manuel Meder for support in setting up simulation environments.
This work was supported through grants of the German Federal Ministry of
Economics and Technology (BMWi, FKZ 50RA1216 and 50RA1217),
the German Federal Ministry for Economic Affairs and Energy
(BMWi, FKZ 50RA1701),
and the European Union's Horizon 2020 research and innovation program
(No H2020-FOF 2016 723853).
}

\bibliographystyle{plain}
\bibliography{literature}

\end{document}